\documentclass[a4paper,twoside]{article}

\usepackage{epsfig}
\usepackage{calc}
\usepackage{amssymb}
\usepackage{amstext}
\usepackage{amsmath}
\usepackage{amsthm}
\usepackage{multicol}
\usepackage{pslatex}
\usepackage[bottom]{footmisc}
\usepackage{algorithmic}
\usepackage{graphicx}
\usepackage{SCITEPRESS}     
\usepackage{float}
\usepackage{algorithm}

\begin{document}

\title{The Effects of Character-Level Data Augmentation on Style-Based Dating of Historical Manuscripts}
\author{\authorname{Lisa Koopmans \orcidAuthor{0000-0001-6556-2600}, Maruf A. Dhali\orcidAuthor{0000-0002-7548-3858} and Lambert Schomaker\orcidAuthor{0000-0003-2351-930X}}
\affiliation{Department of Artificial Intelligence, University of Groningen, The Netherlands}
\email{l.d.koopmans@student.rug.nl, m.a.dhali@rug.nl, L.R.B.Schomaker@rug.nl}
 }

\abstract{Identifying the production dates of historical manuscripts is one of the main goals for paleographers when studying ancient documents. Automatized methods can provide paleographers with objective tools to estimate dates more accurately. Previously, statistical features have been used to date digitized historical manuscripts based on the hypothesis that handwriting styles change over periods. However, the sparse availability of such documents poses a challenge in obtaining robust systems. Hence, the research of this article explores the influence of data augmentation on the dating of historical manuscripts. Linear Support Vector Machines were trained with k-fold cross-validation on textural and grapheme-based features extracted from historical manuscripts of different collections, including the Medieval Paleographical Scale, early Aramaic manuscripts, and the Dead Sea Scrolls. Results show that training models with augmented data improve the performance of historical manuscripts dating by 1\% - 3\% in cumulative scores. Additionally, this indicates further enhancement possibilities by considering models specific to the features and the documents' scripts.}

\onecolumn \maketitle \normalsize \setcounter{footnote}{0} \vfill

\section{\uppercase{Introduction}}\label{sec:introduction}

Handwritten accounts, letters, and similar documents provide essential information about history. To understand such historical manuscripts' social and cultural contexts, paleographers seek to identify their script(s), author(s), location, and production date. Traditionally, paleographers study manuscripts by their writing materials, content, and handwriting styles. However, these methods require specific domain knowledge, are timely processes, and lead to subjective estimations. Additionally, repetitive physical handling leads to further degradation of valuable documents. 

The digitization of historical manuscripts has contributed to their preservation and allowed for the development of automatized methods through machine learning. These tools are more objective than traditional methods and can aid paleographers in assessing their hypotheses. Historical manuscript dating, in particular, can benefit from this, as it can be required to resort to physical methods, which have limited reliability and can be destructive. 

Dates of digitized historical manuscripts have been commonly predicted based on the hypothesis that handwriting styles change over a period \cite{MPS2014}. Thus, manuscripts could be dated by identifying common characteristics in handwriting specific to periods. 

Due to the limited availability of historical manuscripts, research has mainly focused on statistical feature-extraction techniques. These statistical methods extract the handwriting style by capturing attributes such as curvature or slant or representing the general character shapes in the documents \cite{hinge-feature}. However, for reliable results, manuscripts need a sufficient amount of handwriting to extract the handwriting styles. 

Both traditional and automatized methods must deal with data sparsity and the degradation of ancient materials; new data can only be obtained by digitizing or discovering more manuscripts. A possible solution to this issue is data augmentation. Data augmentation is commonly used in machine learning to generate additional realistic training data from existing data to obtain more robust models. However, information on the handwriting styles is lost using standard techniques, such as rotating or mirroring the images. Character-level data augmentation could generate realistic samples simulating an author's variability in handwriting.

Research on the style-based dating of digitized historical manuscripts using data augmentation techniques still needs to be done. Hence, the current research will explore the effects of character-level data augmentation on the style-based dating of digitized historical manuscripts. Manuscript images taken from the Medieval Paleographical Scale (MPS) collections, the Bodleian Libraries of the University of Oxford, the Khalili collections, and the Dead Sea Scrolls were augmented with an elastic rubber-sheet algorithm \cite{bulacu2009recognition}. The first collection, MPS, has medieval charters produced between 1300 and 1550 CE in four cities: Arnhem, Leiden, Leuven, and Groningen. A number of early Aramaic, Aramaic, and Hebrew manuscripts were taken from the last three collections. Several statistical feature-extraction methods on the textural and character level were used to train linear Support Vector Machines (SVM) with only non-augmented images and with both non-augmented and augmented images.

\begin{figure}
        \includegraphics[width=7.5cm]{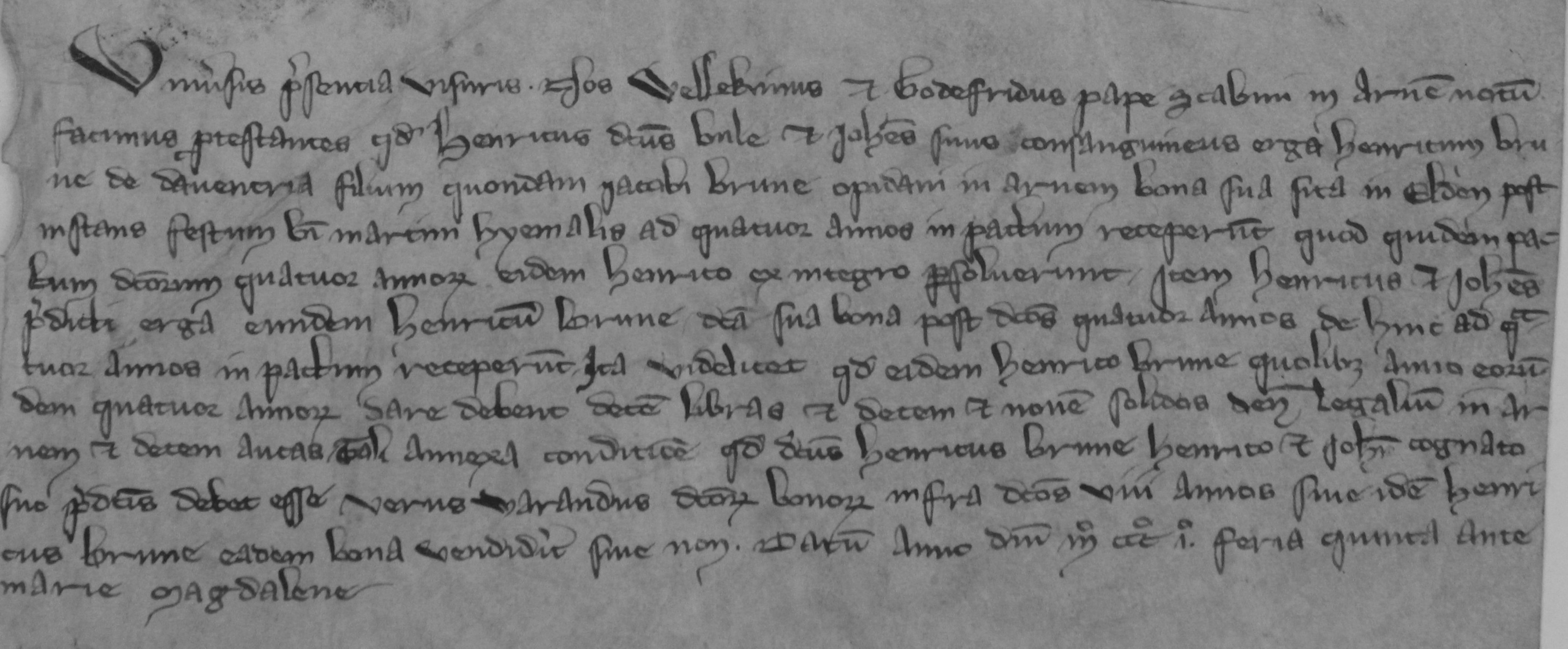}
    \caption{A document image from the Medieval Paleographical Scale (MPS) collection.}
    \label{fig:MPS_ex}
\end{figure}

\section{\uppercase{Related works}}\label{sec:lit_rev}
The main challenge in style-based dating is the selection of feature-extraction techniques. Each script has its own characteristics, which may not be represented well by every feature. Collections of historical manuscripts written in various languages and scripts have been digitized. For example, the Medieval Paleographical Scale \cite{MPSdata} and the Svenskt Diplomatariums huvudkartotek (SDHK) data sets\footnote{https://sok.riksarkivet.se/SDHK} are written in Roman script, consisting of medieval Dutch and Swedish manuscripts respectively. Moreover, the early Aramaic and Dead Sea Scrolls collections \cite{DSSsite} contain ancient texts in Hebrew, Aramaic, Greek, and Arabic, dating from the fifth century BCE (Before the Common Era) until the Crusader Period (12th–13th centuries CE). 

Statistical feature-extraction methods are commonly divided into textural-based features that capture textural information of the handwriting across an entire image and grapheme-based features that capture character-shape information. Graphemes extracted from a set of documents are used to train a clustering method. The cluster representations form a codebook, from which a probability distribution of grapheme usage is computed for each document to represent the handwriting styles.

A widely used textural feature is the 'Hinge' feature, which captures a handwriting sample's slant and curvature information. The features are extensions of the Hinge feature, which describes the joint probability distribution of two hinged edge fragments \cite{hinge-feature}. In addition, Hinge is extended to i.a., co-occurrence features QuadHinge and CoHinge, which emphasize curvature and shape information respectively \cite{Quad+CoHinge-feature}. Other features, such as curvature-free and chain code features, have also been proposed \cite{curvefree}, \cite{TCC}.

Connected Component Contours (CO3) \cite{CO3} is a grapheme-based feature that describes the shape of a fully connected contour fragment. As cursive handwriting has large connected contour fragments, the feature was extended to Fraglets \cite{hinge-feature}, which parts the connected contours based on minima in the fragments. Moreover, k contour fragments (kCF) and k stroke fragments (kSF) features were proposed that partition CO3 in k contour and stroke fragments respectively \cite{kCF-kSF}. Finally, Junclets \cite{Junclets} represents junctions in characters, which are constructed differently in varying writing styles. 

\begin{figure}
    \centering
    \includegraphics{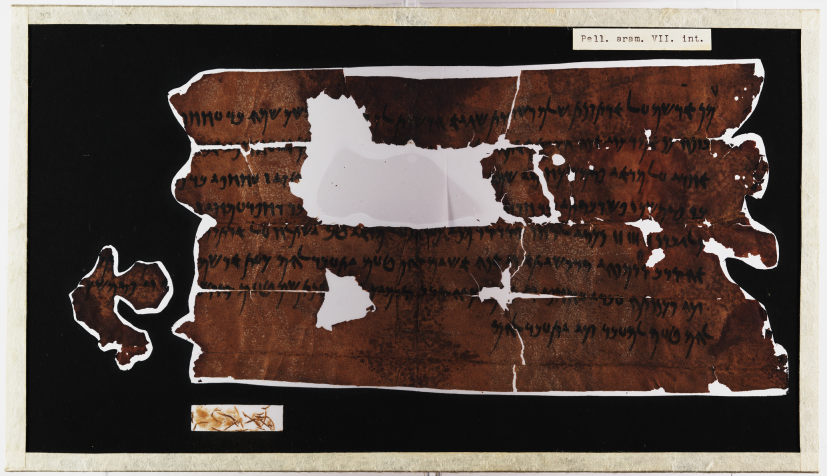}
    \caption{An Early Aramaic (EA) manuscript from the Bodleian Libraries, University of Oxford (Pell. Aram. I).}
    \label{fig:DSS_ex}
\end{figure}

Much research on historical manuscript dating has been done on the MPS data set, specifically by He et al. In \cite{MPS2014}, they predicted dates with a technique combining local and global Support Vector Regression, using Fraglets and Hinge features. They later extended this work, proposing new features such as kCF, kSF, and Junclets. In addition, they proposed the temporal pattern codebook \cite{TPC}, which maintains temporal information lost in the commonly used Self-Organizing Map (SOM) \cite{SOM} to train codebooks. Finally, various statistical feature-extraction methods were compared for historical manuscript dating in \cite{HE2017321}. 

While the MPS data set is relatively clean, it is not representative of many other historical manuscripts. In early works \cite{DHALI2020413}, an initial framework was proposed for the style-based dating of the Dead Sea Scrolls. Unfortunately, the manuscripts from this collection are heavily degraded; many scrolls are fragmented, and ink traces have eroded due to aging. Additionally, the number of labeled manuscripts is small. Therefore, this collection poses a challenge for automatized dating of historical manuscripts. 

\begin{figure}
    \centering
    \includegraphics[width=7.6cm]{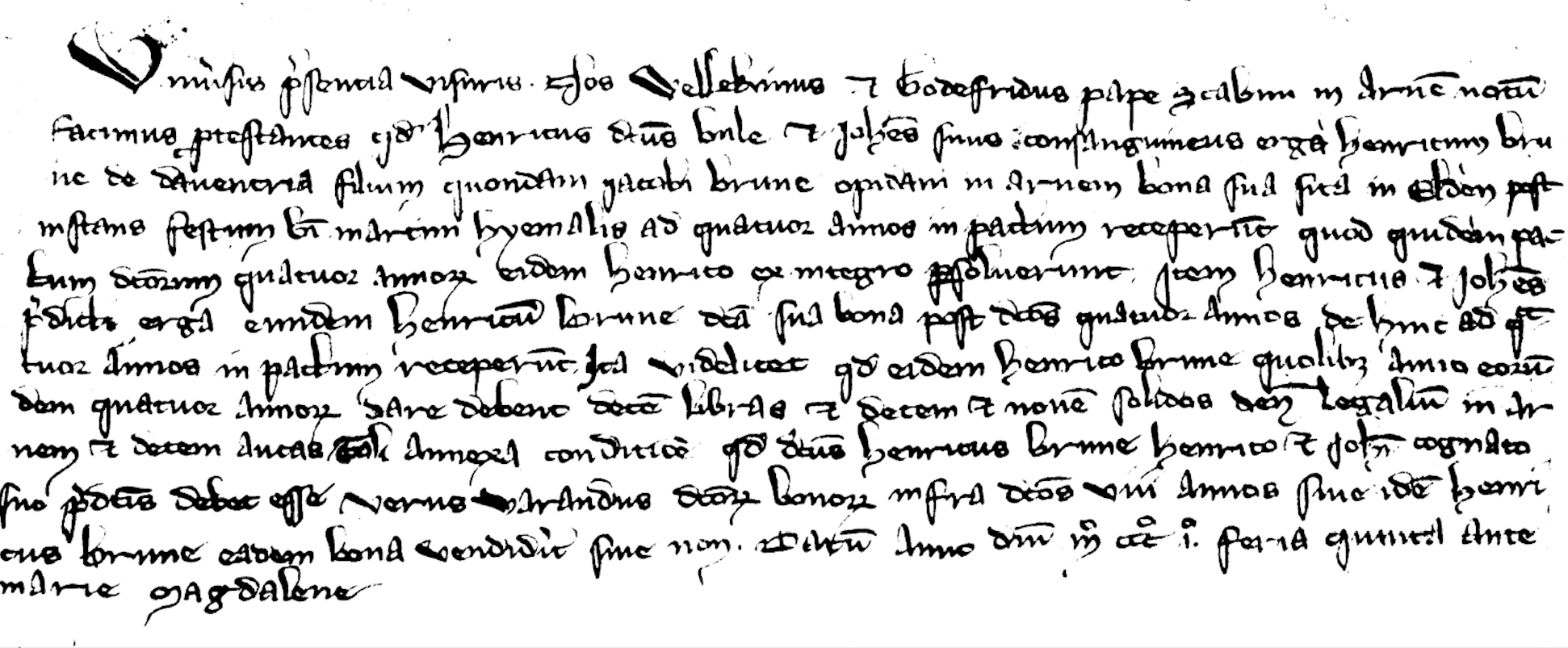}
    \caption{The binarized version of the image from Figure \ref{fig:MPS_ex} with Otsu thresholding.}
    \label{fig:MPS_bin_ex}
\end{figure}

Deep learning approaches have applied transfer learning, meaning pre-trained neural networks were fine-tuned using new data on a different task than initially trained for. This approach requires less data than standard deep learning methods, enabling its use for historical manuscript dating. For example, \cite{CNN2016} used the Google ImageNet-network and fine-tuned it using 11000 images from the SDHK collection. However, this is large for a data set of historical manuscripts. In \cite{CNN2019}, a group of pre-trained neural networks was fine-tuned on the 3267 images from the MPS data set. The best-performing model was shown to outperform statistical methods.

While deep learning approaches show promising results, it is still relevant to consider statistical methods. To train a neural network, the manuscripts' images need to be partitioned into patches, possibly leading to loss of information. To solve this problem, \cite{CNN2019} ensured that each patch contained ``3 to 4 lines of text with 1.5 to 2 words per line'' to extract the handwriting style. While this was a solution for the MPS data set, it may not be for smaller and more degraded collections, such as the Dead Sea Scrolls. In contrast, statistical feature extraction does not require image resizing and considers the handwriting style over the entire image. 

\begin{figure}
    \centering
        \includegraphics[width=7.5cm]{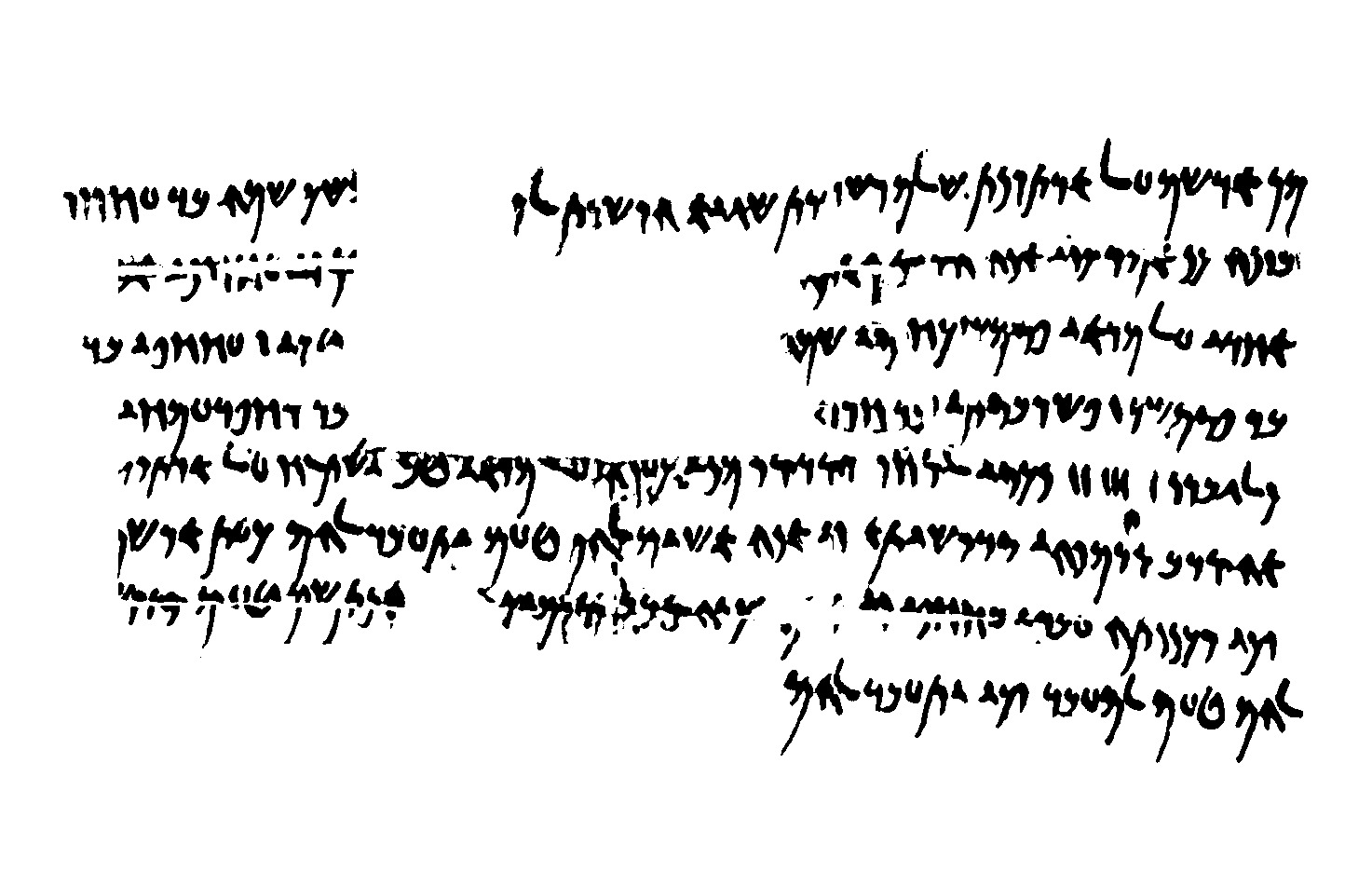}
    \caption{The binarized version of the image from Figure \ref{fig:DSS_ex} using BiNet \cite{BiNet}.}
    \label{fig:DSS_bin_ex}
\end{figure}

\section{\uppercase{Methods}}\label{sec:methods}
This section will present the dating model along with data description, image processing, and feature extraction techniques.
\subsection{Data}\label{sec:_data}
\subsubsection{MPS}
The current research uses the MPS data set \cite{MPS2014},\cite{MPSclustering}, \cite{kCF-kSF}, \cite{MPSdata}. Non-text content, such as seals, supporting backgrounds, color calibrators, etc., have been removed. Consequently, this data set provides relatively clean images. However, some images have been degraded or still contain a small part of a seal or ribbon. The data set is publicly available via Zenodo\footnote{https://zenodo.org/record/1194357\#.YrLU-OxBy3I}.

\begin{table*}[ht]
    \caption{The number of samples over the key years of the MPS data set.}
    \label{_data:chart_per_year}
    \centering
    \begin{tabular}{ c c c c c c c c c c c c}
    \hline
    key year & 1300 & 1325 & 1350 & 1375 & 1400 & 1425 & 1450 & 1475 & 1500 & 1525 & 1550\\
    \hline
    number of charters & 106 & 164 & 199 & 386 & 311 & 323 & 501 & 423 & 372 & 241 & 241\\
    \hline
    \end{tabular}
\end{table*}

The MPS data set contains 3267 images of charters collected from four cities signifying four corners of the medieval Dutch language area. Figure \ref{fig:MPS_ex} shows an example image. Charters were commonly used to document legal or financial transactions or actions. Additionally, their production dates have been recorded. For these charters, usually parchment and sometimes paper was used. 

The charters date from 1300 CE to 1550 CE. Due to the evolution of handwriting being slow and gradual, documents from 11 quarter century key years with a margin of $\pm$ five years were included in the data set. Hence, the data set consists of images of charters from the medieval Dutch language area in the periods 1300 $\pm$ 5, 1325 $\pm$ 5, 1350 $\pm$ 5, up to 1550 $\pm$ 5. Table \ref{_data:chart_per_year} contains the number of charters in each key year.

\subsubsection{Early Aramaic and additional (EAA) manuscripts}
In addition to the MPS data set, 30 images from the early Aramaic, Aramaic, and Hebrew manuscripts were used. For ease of refereeing to this second dataset, EAA is used in the rest of the article, even though EAA contains Aramaic and Hebrew in addition to early Aramaic scripts. A list of the EAA images used in this study can be found in the appendix (see Table \ref{apeen:list}).For these selected manuscripts from the EAA dataset, the dates were directly inferred from dates or events recorded in the manuscripts (i.e., internally dated), and they are publicly available through the Bodleian Libraries, University of Oxford\footnote{https://digital.bodleian.ox.ac.uk/}, the Khalili collections\footnote{https://www.khalilicollections.org/all-collections/aramaic-documents/}, and the Leon Levy Digital Library\footnote{https://www.deadseascrolls.org.il/}. Their dates span from 456 BCE to 133 CE. An example image is shown in Figure \ref{fig:DSS_ex}. In addition, the data set contains several degraded manuscripts with missing ink traces or only two or three lines of text.

\subsection{Preprocessing}\label{sec:_preprocessing}

\subsubsection{Label refinement}
The set of images from the EAA collections did not contain sufficient samples for each year. Therefore, the samples were manually classified based on historical periods identified by historians\footnote{https://www.deadseascrolls.org.il/learn-about-the-scrolls/}. The time periods and the corresponding number of samples are shown in Table \ref{_data:DSS_data}. 

The Persian Period contained two groups of samples spread apart for more than 30 years. Under the speculation that handwriting styles changed during this time, these samples were split into two periods: the Early and Late Persian Periods. These were not based on defined historical periods but on the samples' production years. Images from the upper bound of the year range in Table \ref{_data:DSS_data} were included in the classes. The manuscripts from the Roman Period were excluded as there were insufficient samples. The images were relabeled according to the median of their corresponding year ranges.

\subsubsection{Data augmentation}
To augment the data such that new samples simulate a realistic variability of an author's handwriting, the Imagemorph program \cite{imagemorph} was used. The program applies random elastic rubber-sheet transforms to the data through local non-uniform distortions, meaning that transformations occur on the components of characters. Consequently, the Imagemorph algorithm can generate a large number of unique samples. For the augmented data to be realistic, a smoothing radius of 8 and a displacement factor of 1 were used, measured in units of pixels. As images of the MPS data set required high memory, three augmented images were generated per image. Since the EAA data sets were small, 15 images were generated per image.

\subsubsection{Binarization}
To extract only the handwriting, the ink traces in the images were extracted through binarization. This resulted in images with a white background representing the writing surface, and a black foreground representing the ink of the handwriting. Otsu thresholding \cite{Otsu} was used for binarizing the MPS images, as the MPS data set is relatively clean, and it has been successfully used in previous research with the data set \cite{MPS2014}, \cite{HE2017321}, \cite{kCF-kSF}. Otsu thresholding is an intensity-based thresholding technique where the separability between the resulting gray values (black and white) is maximized. Figure \ref{fig:MPS_bin_ex} shows Figure \ref{fig:MPS_ex} after binarization. 

The EAA images were more difficult to binarize using threshold-based techniques. So, for the EAA images, we used BiNet: a deep learning-based method designed specifically to binarize historical manuscripts \cite{BiNet}. Figure \ref{fig:DSS_bin_ex} shows Figure \ref{fig:DSS_ex} after binarization.

\begin{table*}[ht]
\centering
    \caption{Division of EAA manuscripts across historical time periods. Note that these dates may not exactly be the same as defined by historians.}
    \label{_data:DSS_data}
    \begin{tabular}{l l c c}
    \hline
    Time period & Year range & Median year & Number of samples\\
    \hline
    Early Persian Period & 540 BCE - 400 BCE & 470 BCE & 12\\
    Late Persian Period & 400 BCE - 330 BCE & 365 BCE & 11\\
    Hellenistic Period & 330 BCE - 65 BCE & 198 BCE & 5 \\
    Roman Period & 65 BCE - 325 CE & 195 CE & 2\\
    \hline
    \end{tabular}
\centering
\end{table*}

\subsection{Feature extraction}
The handwriting styles of manuscripts were described by five textural features and one grapheme-based feature. Since the MPS and the EAA data sets are written in different scripts, features were chosen that perform well across different scripts.

\subsubsection{Textural features}

Textural-based feature-extraction methods contain statistical information on handwriting in a binarized image by considering its texture. Textural-based features capture handwriting attributes like slant, curvature, and the author's pen grip, represented in a probability distribution. 

He et al. proposed the joint feature distribution (JFD) principle, describing how new, more robust features can be created \cite{HE2017321}. They identified two groups of such features: the spatial joint feature distribution (JFD-S) and the attribute joint feature distribution (JFD-A). The JFD-S principle derives new features by combining the same feature at adjacent locations, describing a larger area. The JFD-A principle derives new features from different features at the same location and consequently captures various properties. 

\textbf{Hinge} \cite{hinge-feature} is obtained by taking orientations $\alpha$ and $\beta$ with $\alpha < \beta$ of two contour fragments attached at one pixel and computing their joint probability distribution. The Hinge feature captures the curvature and orientation in the handwriting. 23 angle bins were used for $\alpha$ and $\beta$.

\textbf{CoHinge} \cite{Quad+CoHinge-feature} follows the JFD-S principle, combining two Hinge kernels at two different points $x_i, x_j$ with a Manhattan distance $l$, and is described by:
\begin{equation}
    CoHinge(x_i, x_j) = [\alpha_{x_i}, \beta_{x_i}, \alpha_{x_j}, \beta_{x_j}]
\end{equation}
This shows that the CoHinge kernel over contour fragments can be quantized into a 4D histogram. The number of bins for each orientation $\alpha$ and $\beta$ was set to 10.

\textbf{QuadHinge} \cite{Quad+CoHinge-feature} follows the JDF-A principle, combining the Hinge kernel with the fragment curvature measurement $C(f_c)$. Although Hinge also captures curvature information, it focuses on the orientation due to the small lengths of the contour fragments or lengths of the hinge edges. The fragment curvature measurement is defined as:
\begin{equation}
    C(F_c) = \frac{\sqrt{(x_1 - x_2)^2 + (y_1 - y_2)^2}}{s}. 
\end{equation}
$F_c$ is a contour fragment with length $s$ on an ink trace with endpoints $(x_1, y_1), (x_2, y_2)$. In addition, the QuadHinge feature is scale-invariant due to agglomerating the kernel with multiple scales.
The QuadHinge kernel can then be described through the Hinge kernel and the fragment curvature measurement on contour fragments $F_1, F_2$:
\begin{equation}
    H(x_i, s) = [\alpha_{x_i}, \beta_{x_i}, C(F_1), C(F_2)]
\end{equation}
The number of bins of the orientations was set to 12, and that for the curvature to 6, resulting in a dimensionality of $5184$.

\textbf{DeltaHinge} \cite{deltaHinge} is a rotation-invariant feature generalizing the Hinge feature by computing the first derivative of the Hinge kernel over a sequence of pixels along a contour. Consequently, it captures the curvature information of the handwriting contours. The Delta-n-Hinge kernel is defined as:
\begin{equation}
    \begin{cases}
        \Delta ^n \alpha (x_i) = \frac{\Delta ^{n-1} \alpha (x_i) - \Delta ^{n-1}\alpha (x_i + \delta l)}{\delta l}\\
        
         \Delta ^n \beta (x_i) = \frac{\Delta ^{n-1} \beta (x_i) - \Delta ^{n-1}\beta (x_i + \delta l)}{\delta l}
    \end{cases}
\end{equation}
Where $n$ is the $nth$ derivative of the Hinge kernel. When used for writer identification, performance decreased for $n >1$, implying that the feature's ability to capture writing styles decreased. Hence, the current research used $n = 1$. 

\textbf{Triple Chain Code (TCC)} \cite{TCC} captures the curvature and orientation of the handwriting by combining chain codes at three different locations along a contour fragment. The chain code represents the direction of the next pixel, indicated by a number between 1 to 8. TCC is defined as:
\begin{equation}
    TCC(x_i, x_{i+l}, x_{i + 2l}) = [CC(x_i), CC(x_{i+l}), CC(x_{i+2l})]
\end{equation}
Where $CC(x_i)$ is the chain code at location $x_i$, and Manhattan distance $l = 7$.

\subsubsection{Grapheme-based Features}
Grapheme-based features are allograph-level features that partially or fully overlap with allographs in handwriting, described by a statistical distribution. The handwriting style is then represented by the probability distribution of the grapheme usage over a document, computed with a common codebook.

\textbf{Junclets} \cite{Junclets} represents the crossing points, i.e., junctions, in handwriting. Junctions are categorized into `L`, `T`, and `X` junctions with 2, 3, and 4 branches, respectively. In different time periods, the angles between the branches, the number of branches, and the lengths of the branches can differ, making the feature appropriate for dating. Compared to other grapheme-based features, this feature does not need segmentation or line detection methods. A junction is represented as the normalized stroke-length distribution of a reference point in the ink over a set of $N = 120$ directions. The stroke lengths are computed with the Euclidean distance from a reference point in a direction until the edge of the ink. The feature is scale-invariant and captures the ink-width and stroke length.

\subsubsection{Codebook}
Previous research commonly used the Self-Organizing Map (SOM) \cite{SOM} unsupervised clustering method to train the codebook \cite{HE2017321}. By using this, however, temporal information in the input patterns is lost. The partially supervised Self-Organizing Time Map (SOTM) \cite{SOTM} maintains this information. In \cite{TPC}, SOTM showed an improved performance for a grapheme-based feature compared to SOM. Hence, the codebook was trained with SOTM. 

SOTM trains sub-codebooks $D_t$ for each time period using the standard SOM \cite{SOM}, with handwriting patterns $\Omega (t)$ from key year $y(t)$. The key years for the MPS (in CE) and the EAA (in BCE) data sets were defined as $y(t) = \{1300, 1325, 1350, ..., 1550\}$, and $y(t) = \{470, 365, 198 \}$ respectively. The final codebook $D$, is composed of the sub-codebooks $D_t$: $D = \{D_1, D_2, ..., D_n\}$, with $n$ key years. To maintain the temporal information, the sub-codebooks are trained in ascending order. The initial sub-codebook $D_1$ is randomly initialized as no prior information exists in the data set. The succeeding sub-codebooks are initialized with $D_{t - 1}$ and then trained. Algorithm \ref{alg:SOTM} shows the pseudo-code obtained from \cite{TPC}. 

To train the sub-codebooks, the Euclidean distance measure was used as it significantly decreased training times. Each sub-codebook was trained for 500 epochs to ensure sufficient training took place. The learning rate $\alpha^*$ decayed from $\alpha = 0.99$ following \eqref{eq:sotm_decay}. The sub-codebooks were trained on a computer cluster \footnote{https://wiki.hpc.rug.nl/peregrine/start}. 
\begin{equation}
    \label{eq:sotm_decay}
    \alpha^* = \alpha \cdot \left( 1 - \frac{current \ epoch}{max \ epoch}\right)
\end{equation}

A historical manuscript's feature vector was obtained by mapping its extracted graphemes to their most similar elements in the trained codebook, computed via the Euclidean distance, and forming a histogram. Finally, the normalized histogram formed the feature vector.

\begin{algorithm}[ht]
    \caption{SOTM \cite{TPC}}
    \label{alg:SOTM}
    \begin{algorithmic}
    \STATE $t \Leftarrow 1$
    \STATE Randomly initialize $D_t$
    \STATE Train $D_t$ using $\Omega (t)$ by the standard SOM
    \WHILE{$t \leq n$}
    \STATE $t \Leftarrow t + 1$
    \STATE Initialise $D_t$ using $D_{t - 1}$
    \STATE Train $D_t$ using $\Omega (t)$ by the standard SOM
    \ENDWHILE
    \STATE Output $D = \{D_1, D_2, ..., D_t, ..., D_n\}$
    \end{algorithmic}
\end{algorithm}

\subsection{Post-processing}
The feature vectors of all features were small decimal numbers, varying between $10^{-2}$ and $10^{-6}$. To emphasize the differences between the feature vectors of a type of feature, the feature vectors were normalized between 0 and 1 based on the range of a feature's feature vectors. A feature vector $f$ is scaled according to the following equations:

\begin{equation}
    f_{std} = \frac{f - min(f)}{max(f) - min(f)}
\end{equation}
\begin{equation}
    f_{scaled} = f_{std} \cdot (max - min) + min \ \
\end{equation}

Here, $max$ and $min$ are the maximum and minimum values over the whole set of feature vectors of a certain feature, while $max(f)$ and $min(f)$ are the maximum and minimum values of the feature vector $f$ \cite{scikit-learn}. 

\subsection{Dating}
\subsubsection{Model}
Historical manuscript dating can be regarded as a classification or a regression problem. As the MPS data set was divided into 11 classes (or key years) with clear borders, and the EAA data set was partitioned into classes, it was regarded as a classification problem. Following previous research on the MPS data set \cite{HE2017321}, linear Support Vector Machines (SVM) were used for date prediction with a one-versus-all strategy. 

\subsubsection{Measures}
The Mean Absolute Error (MAE) and the Cumulative Score (CS) are two commonly used metrics to evaluate model performance for historical manuscript dating. The MAE is defined as follows:
\begin{equation}
    MAE = \frac{\sum_{i=0}^{N-1}  \vert y_i - \Bar{y_i} \rvert}{N}
\end{equation}
Here, $y_i$ is a query document's ground truth, and $\Bar{y_i}$ is its estimated year. $N$ is the number of test documents. The CS is defined in \cite{CS} as
\begin{equation}
    CS = \frac{N_{e <= \alpha}}{N} \cdot 100 \%
\end{equation}
The CS describes the percentage of test images that are predicted with an absolute error $e$ no higher than a number of years $\alpha$. At $\alpha = 0$ years, the CS is equal to the accuracy. 

For both the MPS and the EAA data sets, CS with $\alpha = 0$ years was used. Since paleographers generally consider an absolute error of 25 years acceptable, and the MPS set has key years spread apart by 25 years, CS with $\alpha = 25$ years was also used for this data set.

\begin{figure}
    \centering
    \includegraphics[width=7.5cm]{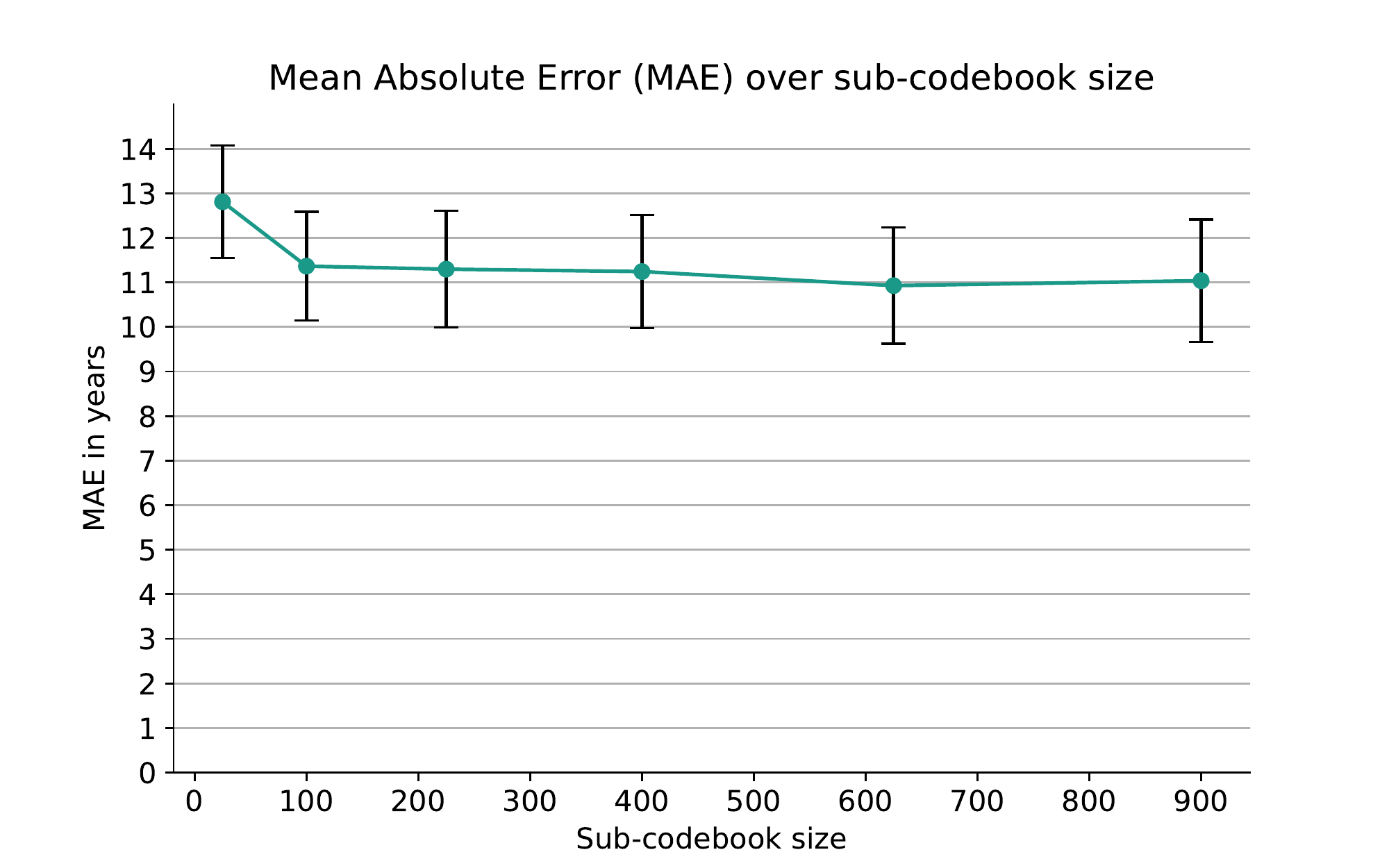}
    \caption{MAE over sub-codebook size on non-augmented MPS data from 10-fold cross-validation.}
    \label{fig:MPS_codebook_mae}
\end{figure}

\begin{figure}
    \centering
    \includegraphics[width=7.5cm]{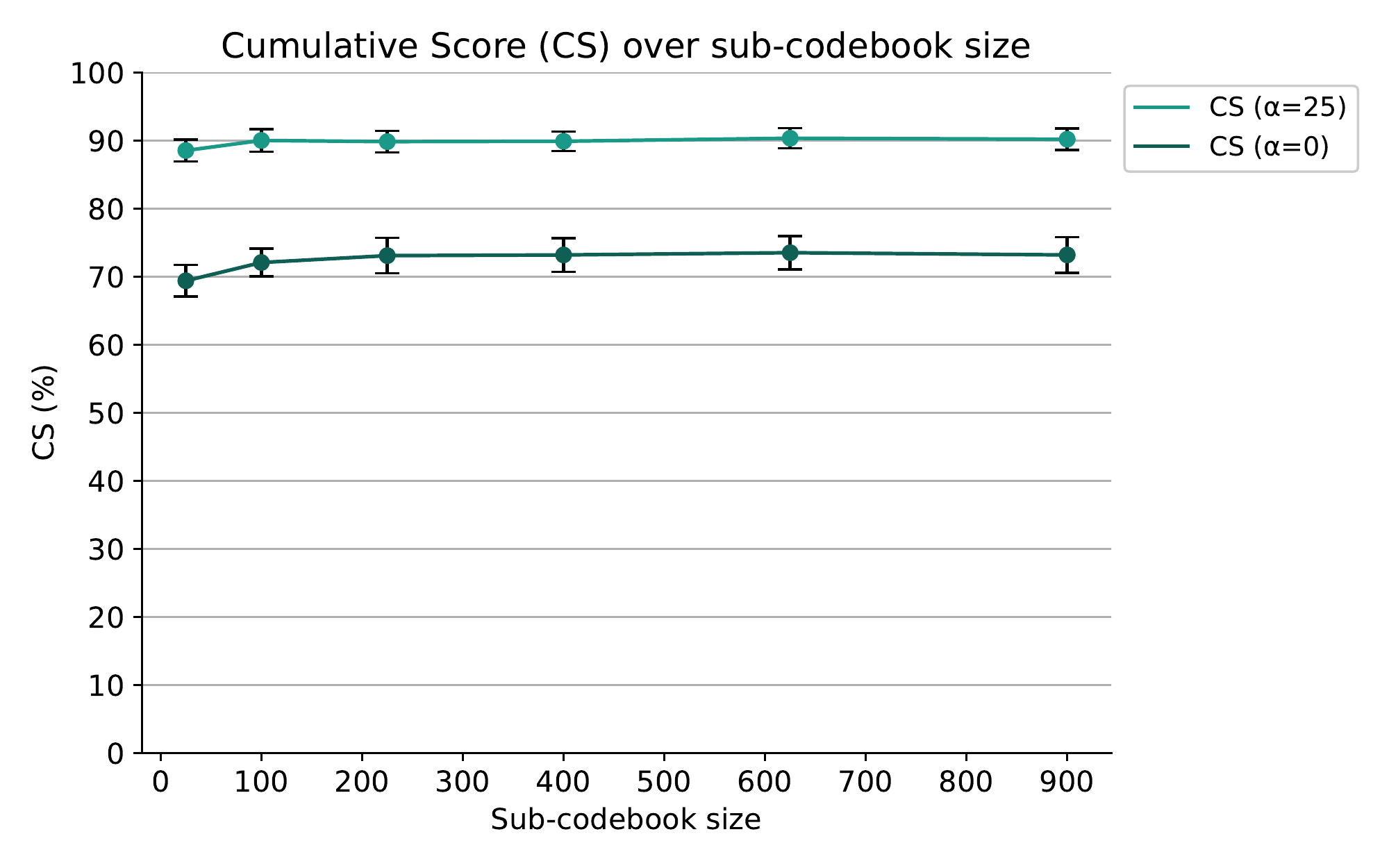}
    \caption{CS with $\alpha = 25$ and $\alpha = 0$ years over sub-codebook size on non-augmented MPS data from 10-fold cross-validation.}
    \label{fig:MPS_codebook_cs}
\end{figure}

\begin{figure}
    \centering
    \includegraphics[width=7.5cm]{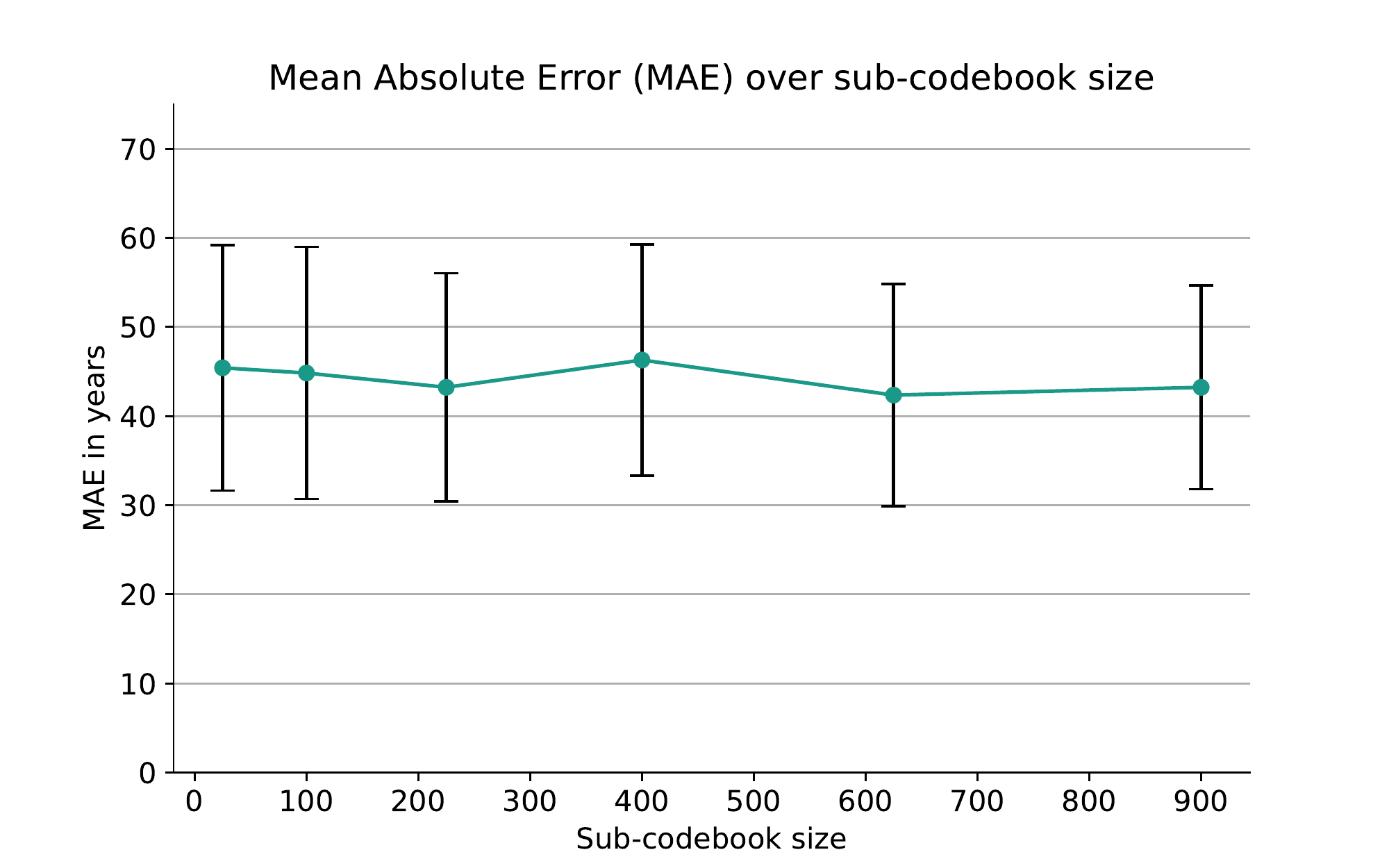}
    \caption{MAE over sub-codebook size on non-augmented EAA data from 4-fold cross-validation.}
    \label{fig:DSS_codebook_mae}
\end{figure}

\begin{figure}
    \centering
    \includegraphics[width=7.5cm]{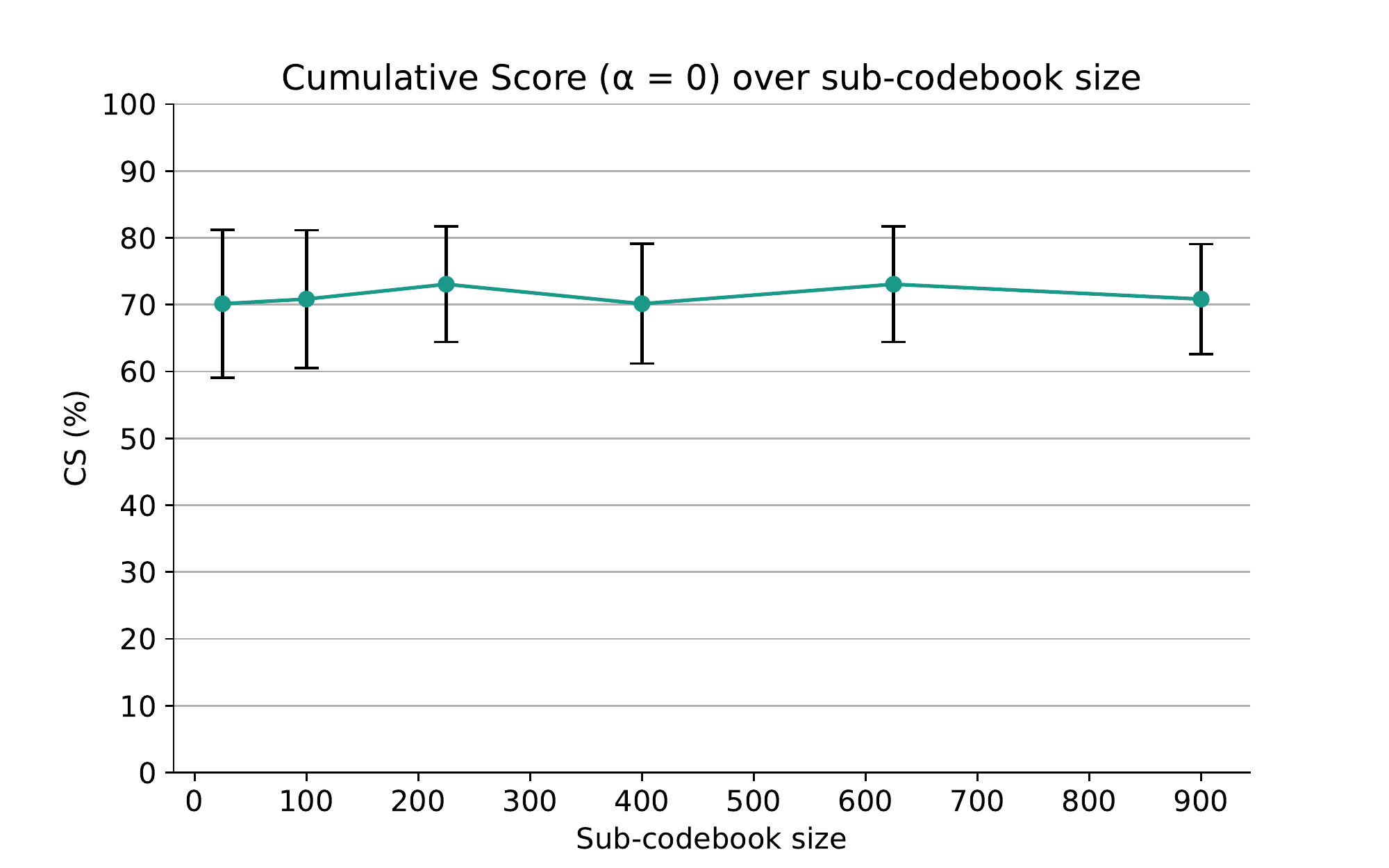}
    \caption{CS with $\alpha = 0$ years over sub-codebook size on non-augmented EAA data from 4-fold cross-validation.}
    \label{fig:DSS_codebook_cs}
\end{figure}

\subsubsection{Experiments}
The MPS images were randomly split into a test and training set, containing 10\% and 90\% of the data, respectively. The EAA images were split into a test set of 5 images and a training set of 23 images. 2 samples were included from classes 470 and 365 BCE each. As the class 198 BCE contained only five images, one image from this class was considered in the test set. The images were sorted based on their labels, and the first images of each class were selected for testing. 

The models were tuned with stratified k-fold cross-validation for both data sets, as they were imbalanced. For the MPS data set, $k = 10$. Since the training set of the EAA data set contained only four images from  198 BCE, $k = 4$ for this set. To prevent a randomized split in each iteration of the k-fold cross-validation from affecting the selection of hyper-parameters, hyper-parameters were selected using the mean results of stratified k-fold cross-validation across six random seeds, ranging from 0 to 250 with steps of 50. The set of values considered for the hyper-parameters were $2^n, n = -7, -6, -5, ..., 10$. During the process, the augmented images of those in the validation and test sets were excluded from the training sets.

Models were trained in two conditions. In the non-augmented condition only non-augmented images were used, and in the augmented condition both augmented and non-augmented images were used for training.

\textbf{Codebook}   Different sub-codebook sizes can result in different model performances. Hence, various sub-codebook sizes were tested to obtain the size for the Junclets feature. A codebook's size is its total number of nodes, i.e., $n_{columns} \cdot n_{rows}$. The full codebook $D$ is the concatenation of the sub-codebooks $D_t$, and thus its size will be $size_{D_t} \cdot n_{classes}$. The set of sub-codebook sizes $s = \{25, 100, 225, 400, 625, 900\}$ with $n_{columns} = n_{rows}$ were considered. These conditions were the same for the MPS and the EAA images. Since different codebook sizes result in different features, the sub-codebook sizes were determined based on the validation results of models trained on only non-augmented images. 

The code used for the experiments and the SOTM is publicly available\footnote{https://github.com/Lisa-dk/Bachelor-s-thesis.git}.

\section{\uppercase{Results}}\label{sec:results}
To explore the effects of data augmentation on the style-based dating of historical manuscripts,  five textural features and one grapheme-based feature were used. Linear SVMs were trained using only non-augmented data in the 'non-augmented' condition, and using both augmented and non-augmented data in the 'augmented' condition. The models were tuned with stratified 10-fold (MPS) and 4-fold (EA) cross-validation and tested on a hold-out set containing only non-augmented data. The test set of the MPS data set contained 10\% of the data, and that of the EAA dataset contained 17.8\% (5 images) of the data. 

The models were evaluated with the MAE and CS with $\alpha = 0$ years (i.e. accuracy). In addition, the MPS data set was also evaluated with CS with $\alpha = 25$ years.

\subsection{Sub-codebook size}
To investigate Junclets, first, an optimal sub-codebook size needed to be selected. Results of k-fold cross-validation for sub-codebook sizes 25, 100, 225, 400, 625, and 900 were evaluated on non-augmented data. 

Figures \ref{fig:MPS_codebook_mae} and \ref{fig:MPS_codebook_cs} show the MAE and CS for the MPS data set over sub-codebook size, respectively. The MAE shows a minimum at the sub-codebook size of 625. Moreover, CS with $\alpha = 25$ and $\alpha = 0$ years show a maximum at sub-codebook size 625. Therefore, Junclets features were obtained with sub-codebooks of size 625 on the MPS data.

Figure \ref{fig:DSS_codebook_mae} displays the MAE over the sub-codebook size on validation results for the EAA data. The MAE decreases until the sub-codebook size is 225, after which it fluctuates. This is reflected in the CS with $\alpha = 0$ years (Figure \ref{fig:DSS_codebook_cs}), which displays an increase until size 225, after which it fluctuates. In addition, the standard deviations for the MAE and CS ($\alpha = 0$) appear the smallest here. Hence, a sub-codebook size of 225 was chosen for the EAA data.

\begin{figure}[ht]
    \centering
    \includegraphics[width=7.5cm]{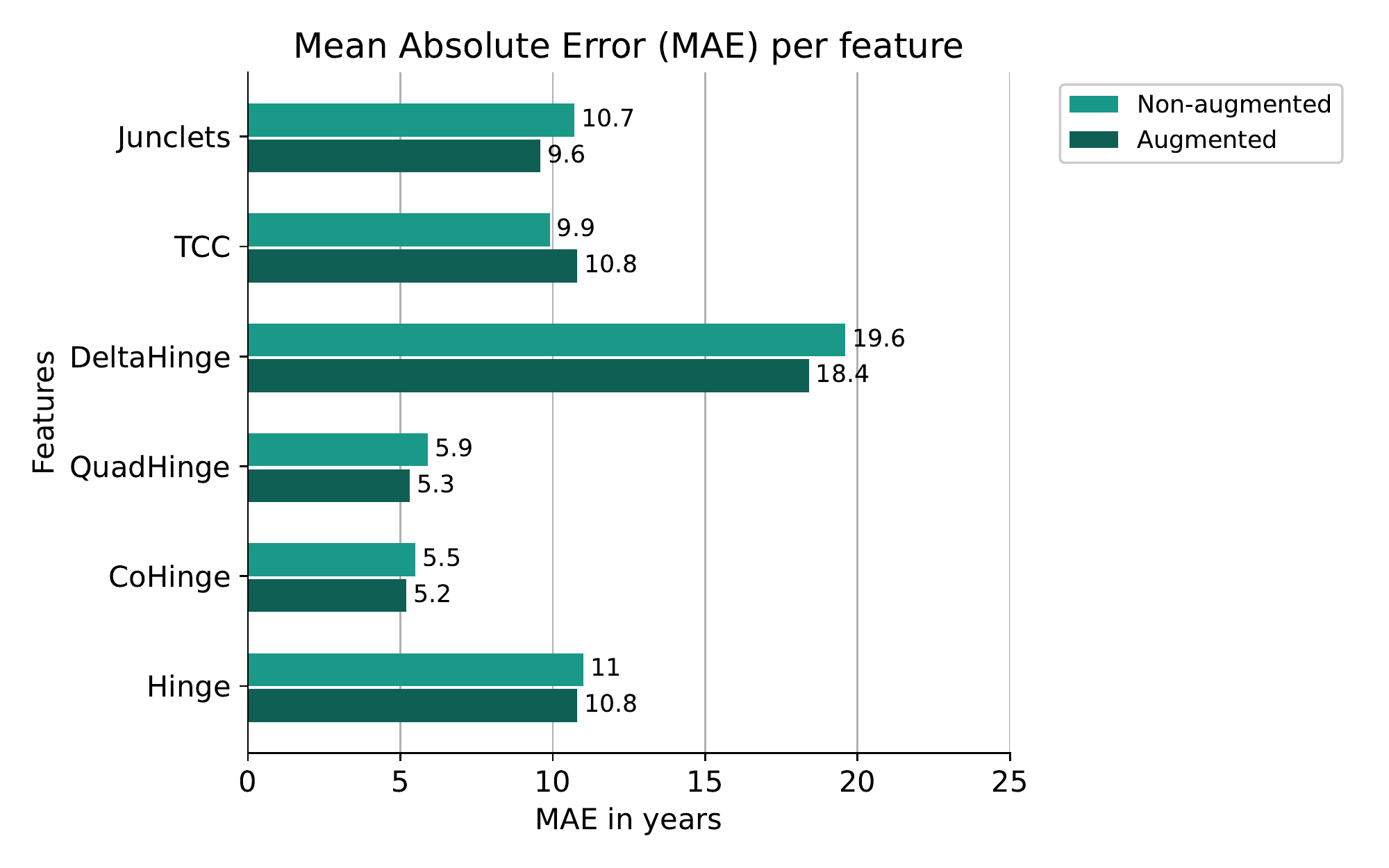}
    \caption{MAE on MPS (unseen) test data across non-augmented and augmented conditions.}
    \label{fig:MPS_test_mae}
\end{figure}

\begin{figure}[ht]
    \centering
    \includegraphics[width=7.5cm]{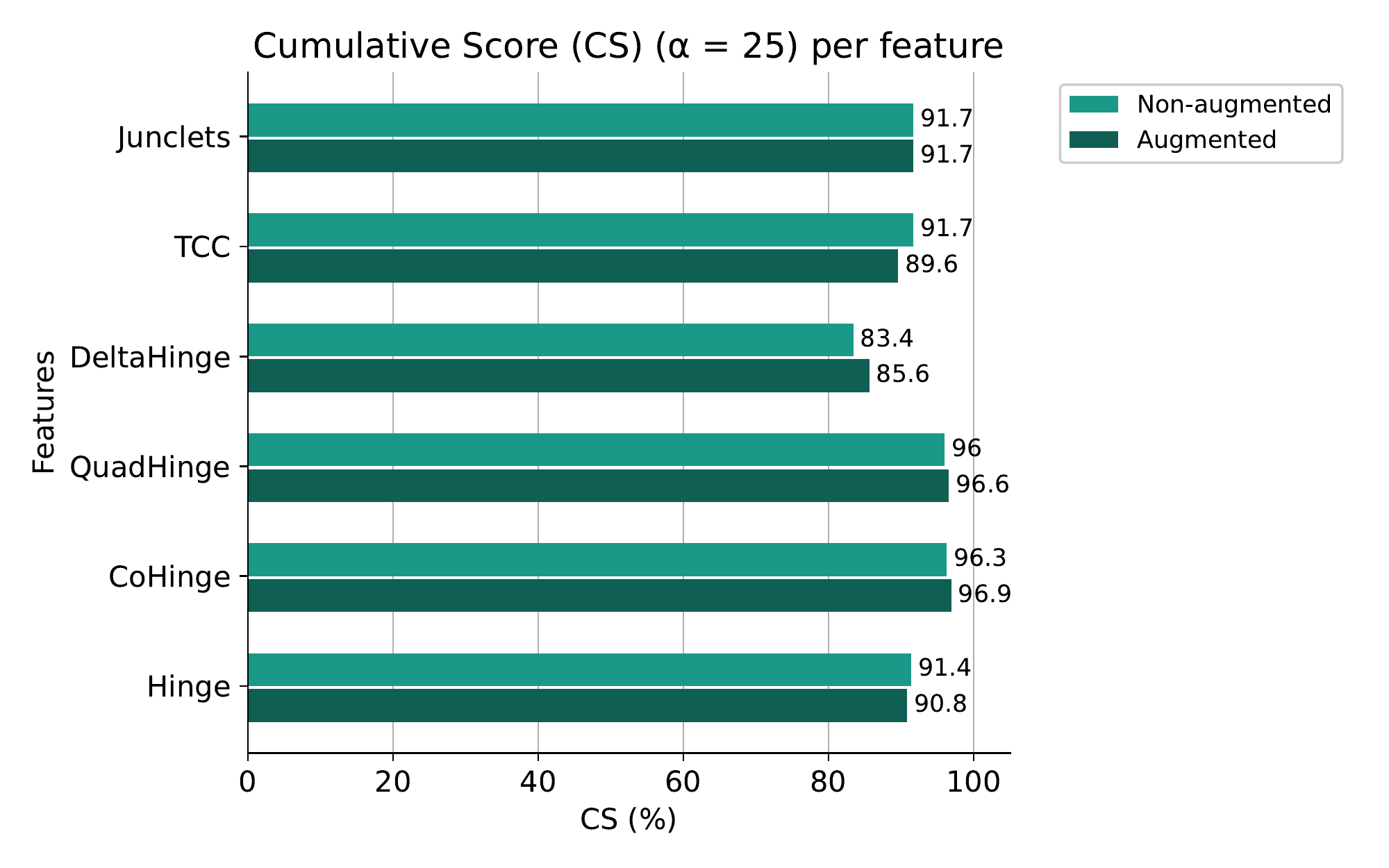}
    \caption{CS with $\alpha = 25$ years on MPS (unseen) test data across non-augmented and augmented conditions.}
    \label{fig:MPS_test_cs_25}
\end{figure}

\begin{figure}[ht]
    \centering
    \includegraphics[width=7.5cm]{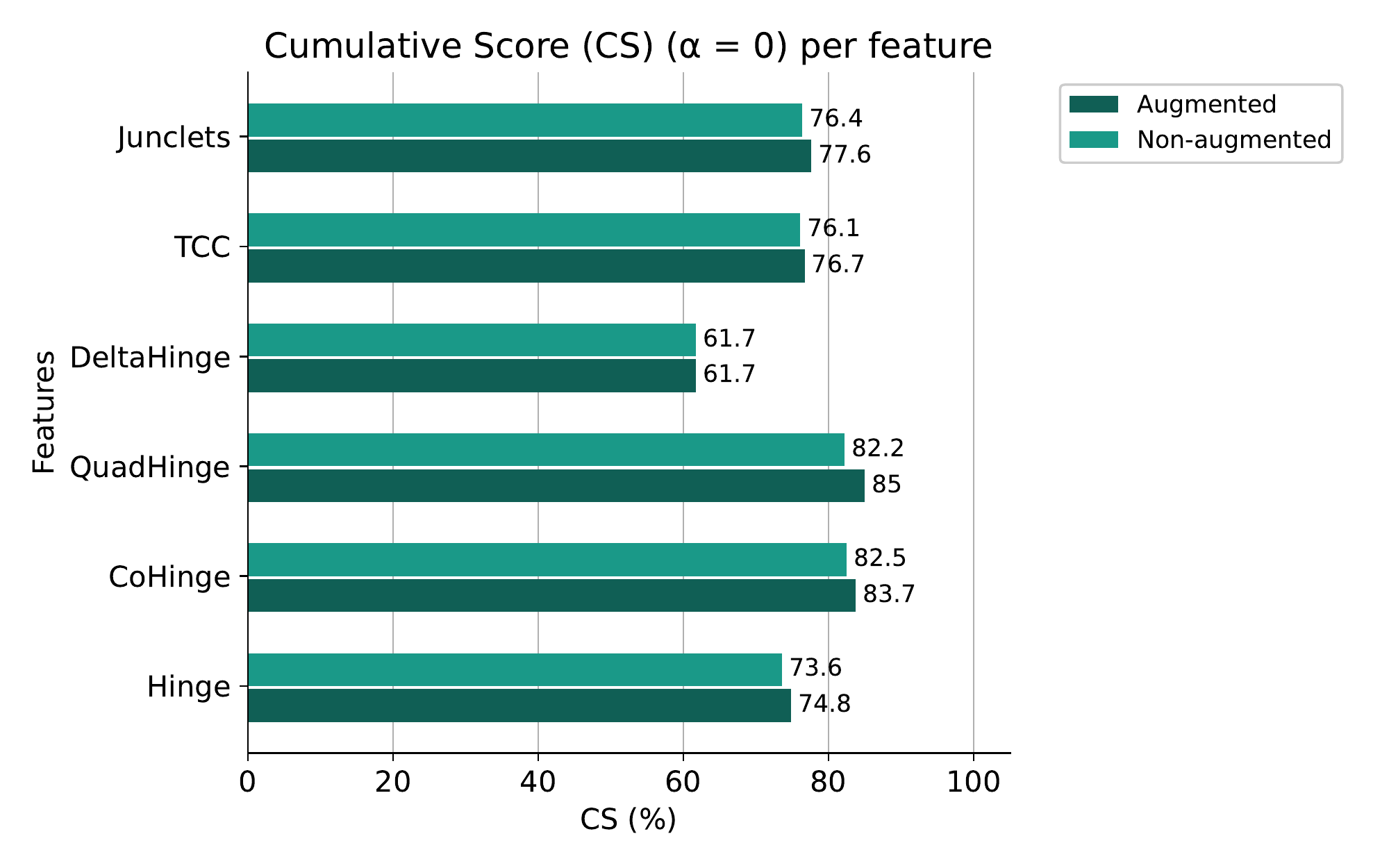}
    \caption{CS with $\alpha = 0$ years on MPS (unseen) test data across non-augmented and augmented conditions.}
    \label{fig:MPS_test_cs_0}
\end{figure}

\begin{table*}[ht]
    \caption{k-fold cross-validation results on the MPS data set.}
    \centering
    \label{_data:MPS_cv}
    \scalebox{0.8}{\centering
    \begin{tabular}{l l l l l l l l}
    \hline
    & Non-augmented & Augmented & Non-augmented & Augmented & Non-augmented & Augmented\\
    Feature & MAE & MAE &  CS ($\alpha$=25) & CS ($\alpha$=25) & CS ($\alpha$=0) & CS ($\alpha$=0)\\
    \hline
    Junclets   & 10.93 $\pm$ 1.31 & 9.15 $\pm$ 1.29 & 90.35 $\pm$ 1.49 & 92.39 $\pm$ 1.51 & 73.53 $\pm$ 2.45 & 77.56 $\pm$ 2.31 \\
    TCC        & 9.47 $\pm$ 1.08 & 8.95 $\pm$ 1.17 & 91.37 $\pm$ 1.39 & 92.16 $\pm$ 1.47 & 77.00 $\pm$ 1.85 & 77.98 $\pm$ 2.10\\
    DeltaHinge & 20.08 $\pm$ 1.88 & 18.35 $\pm$ 1.55 & 81.59 $\pm$ 1.96 & 83.00 $\pm$ 1.78 & 61.60 $\pm$ 2.25 & 63.91 $\pm$ 2.10\\
    QuadHinge  & 5.76 $\pm$ 0.97 & 5.74 $\pm$ 0.97 & 95.38 $\pm$ 1.16 & 95.44 $\pm$ 1.17 & 84.65 $\pm$ 1.89 & 84.53 $\pm$ 1.94 \\
    CoHinge    & 6.81 $\pm$ 0.96 & 6.48 $\pm$ 0.88 & 94.32 $\pm$ 1.23 & 94.59 $\pm$ 1.17 & 82.13 $\pm$ 1.93 & 82.64 $\pm$ 1.95\\
    Hinge      & 11.55 $\pm$ 1.44 & 11.28 $\pm$ 1.38 & 89.42 $\pm$ 1.74 & 89.36 $\pm$ 1.76 & 73.60 $\pm$ 2.55 & 73.76 $\pm$ 2.52 \\
    \hline
    \end{tabular}
    }
\end{table*}

\begin{table*}[ht]
    \caption{k-fold cross-validation results on the EAA data set.}
    \label{_data:DSS_cv}
    \centering
    \scalebox{0.8}{
    \begin{tabular}{l l l l l l}
    \hline
    & Non-augmented & Augmented & Non-augmented & Augmented\\
    Feature & MAE & MAE & CS ($\alpha$=0) & CS ($\alpha$=0)\\
    \hline
    Junclets  & 43.22 $\pm$ 12.80 & 40.40 $\pm$ 20.78 & 73.05 $\pm$ 8.66 & 70.28 $\pm$ 13.68 \\
    TCC & 47.26 $\pm$ 12.52 & 57.92 $\pm$ 18.77 & 71.94 $\pm$ 10.59 & 65.42 $\pm$ 12.53 \\
    DeltaHinge & 46.72 $\pm$ 9.13 & 45.54 $\pm$ 22.79 & 65.83 $\pm$ 8.20 & 75.55 $\pm$ 12.80 \\
    QuadHinge & 38.97 $\pm$ 13.53 & 29.92 $\pm$ 15.72 & 76.67 $\pm$ 8.10 & 82.08 $\pm$ 9.41 \\
    CoHinge & 48.18 $\pm$ 8.74 & 38.95 $\pm$ 20.93 & 64.44 $\pm$ 7.97 & 75.28 $\pm$ 12.69 \\
    Hinge & 33.84 $\pm$ 13.43 & 26.17 $\pm$ 20.63 & 79.86 $\pm$ 7.07 & 84.44 $\pm$ 11.28 \\
    \hline
    \end{tabular}
    }
\end{table*}

\subsection{Augmentation}
\subsubsection{MPS}
Figure \ref{fig:MPS_test_mae} shows the MAE for each feature across the augmented and non-augmented conditions. The MAE for TCC increased in the augmented condition compared to the non-augmented condition. All other features displayed a decrease in the augmented condition.

Figure \ref{fig:MPS_test_cs_25} shows the CS with $\alpha = 25$ years for both non-augmented and augmented conditions. An increase occurred in the augmented condition compared to the non-augmented condition for all features, except for TCC and Hinge, which display a decrease. Additionally, Junclets did not change in performance across conditions.

As displayed in Figure \ref{fig:MPS_test_cs_0}, all features showed an increase in CS with $\alpha = 0$ years in the augmented condition compared to the non-augmented condition with the exception of DeltaHinge. This feature showed no change in performance on test data.

These results denote an overall increase in performance for all features, with the exception of TCC. However, the changes in performances are small, which is reflected in the validation results displayed in Table \ref{_data:MPS_cv}, where changes between the non-augmented and augmented conditions are insignificant. This is indicated by means of the measures in augmented conditions falling within the ranges denoted by the standard deviations of the non-augmented conditions.

\begin{figure}[ht]
    \centering
    \includegraphics[width=7.5cm]{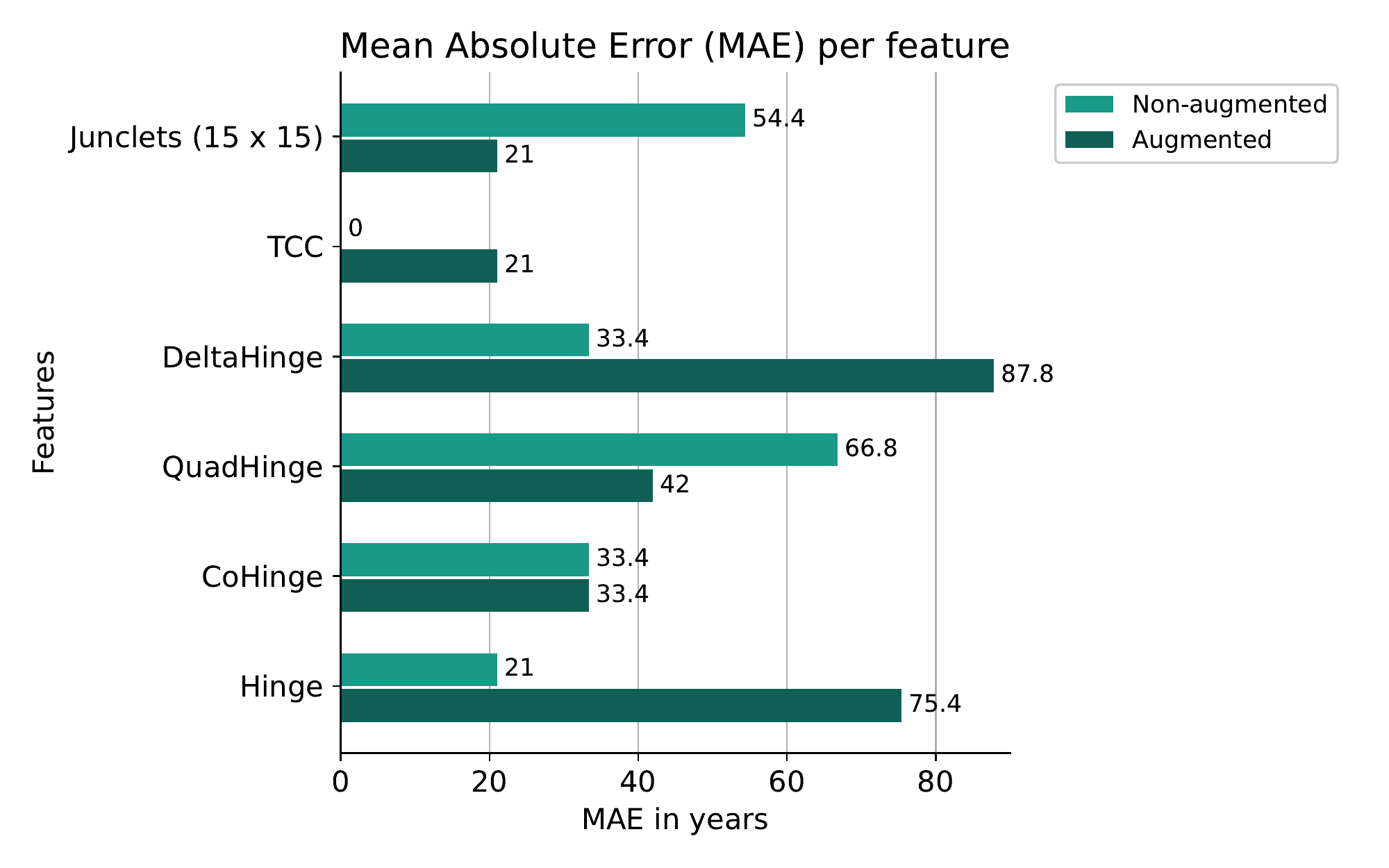}
    \caption{MAE on EAA (unseen) test data across non-augmented and augmented conditions.}
    \label{fig:DSS_test_mae}
\end{figure}

\begin{figure}[ht]
    \centering
    \includegraphics[width=7.5cm]{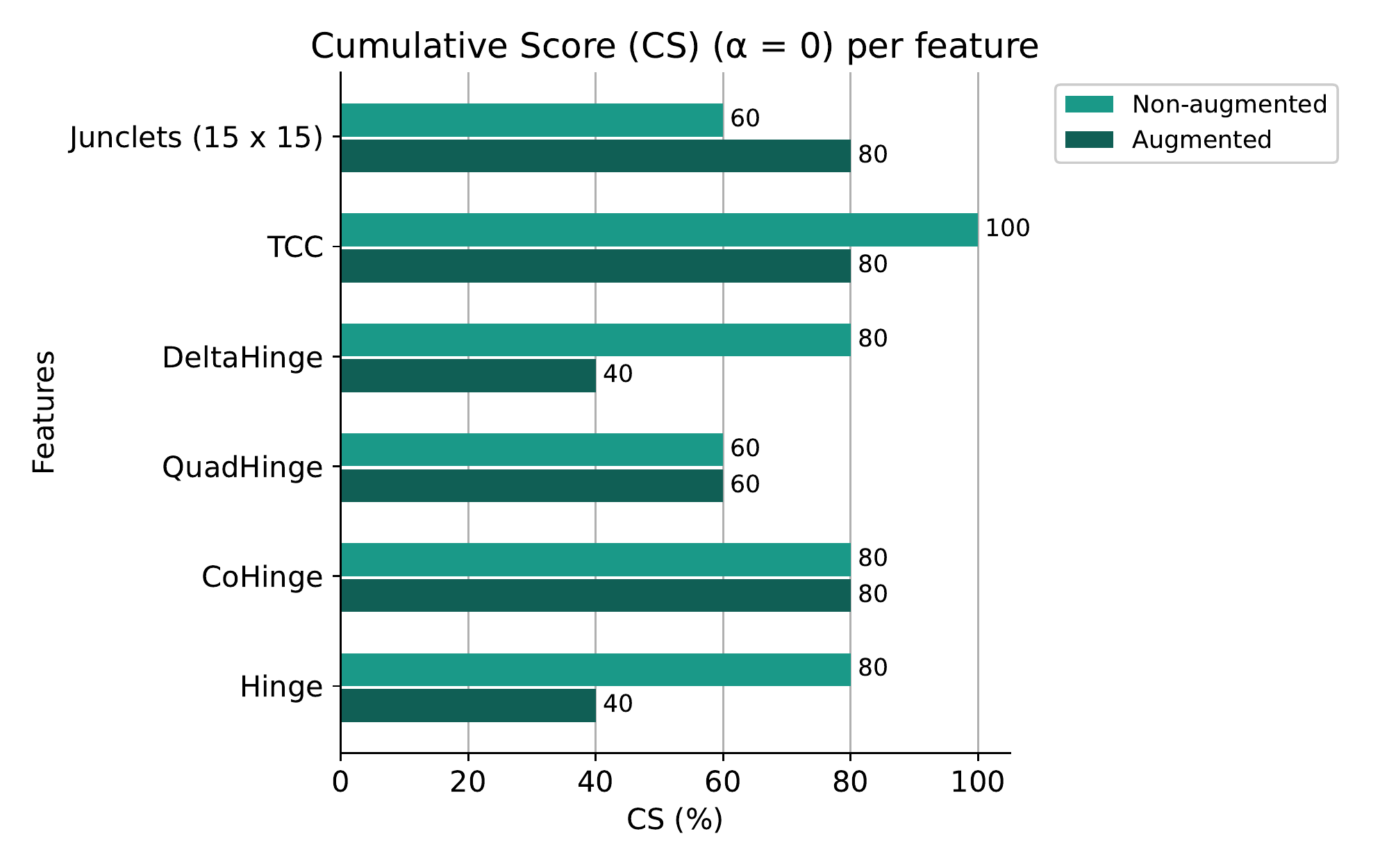}
    \caption{CS with $\alpha = 0$ years on EAA (unseen) test data across non-augmented and augmented conditions.}
    \label{fig:DSS_test_cs_0}
\end{figure}

\subsubsection{EAA Collections}
Figures \ref{fig:DSS_test_mae} and \ref{fig:DSS_test_cs_0} show the MAE and CS with $\alpha = 0$ years across all features for the EAA data set. Performance increased for Junclets in the augmented condition compared to the non-augmented condition, indicated by the decrease in MAE and increase in accuracy. QuadHinge also showed an increase in performance as the MAE decreased in the augmented condition. A decrease in performance for TCC, DeltaHinge, and Hinge features is denoted by an increase in MAE and a reduction in accuracy. CoHinge displayed no change across conditions.

These results are not reflected in the validation results (Table \ref{_data:DSS_cv}), where Junclets and TCC displayed a decrease in performance with a reduction in mean MAE and an increase in mean accuracy in the augmented condition compared to the non-augmented condition. DeltaHinge, QuadHinge, CoHinge, and Hinge, however, displayed the opposite. Additionally, standard deviations increased significantly in the augmented condition compared to the non-augmented condition.

\subsubsection{Significance}
A statistical test (ANOVA, \cite{cuevas2004anova}) was performed to see if the results showed significant improvement. For the MPS data, the results from Junclets feature were statistically significant for both MAE and CS, with p-values much smaller than $0.005$. However, for the EAA data, it did not show any significance for any of the feature extraction techniques.

\section{\uppercase{Discussion}}\label{sec:discussion}
The current study explores the effects of character-level data augmentation on the style-based dating of historical manuscripts using images from the MPS and EAA collections. Images were augmented with the Imagemorph program \cite{imagemorph} and then binarized. Linear SVMs were trained on five textural features and one grapheme-based feature. The grapheme-based feature Junclets was obtained by mapping extracted junction representations to a codebook trained with SOTM \cite{SOTM}. Experiments were conducted to determine the sub-codebook sizes. SVMs were trained in `non-augmented' and `augmented' conditions where only non-augmented images and both non-augmented and augmented images were used, respectively. Models were evaluated through the MAE and CS with $\alpha$-values of 0 and 25 years.

\subsection{Key findings}
\subsubsection{MPS}
Test results showed that linear SVMs trained on MPS data in the augmented condition displayed an overall increased performance compared to the non-augmented condition for all features except TCC. TCC showed a decrease in performance. However, these increases and decreases were small, and changes in validation results were insignificant, with the ranges of the standard deviations and means overlapping across conditions. 

The MPS images require much computer memory and, consequently, long running times to acquire the features and models. Specifically, obtaining the Junclets features required several days. Hence, only three augmented images per MPS image were generated. Were more images generated, results might have shown a clearer picture of the influence of data augmentation on historical manuscript dating. 

Another possible explanation for the small changes in performance shown by the MPS data set results is that MPS images were augmented before binarization. The Imagemorph program applies a Gaussian filter over local transformations. Consequently, if it is applied before binarization, the background's influence leads to less severe distortions than if it is applied after binarization. Although the distortions were noticeable, they might have been too light to produce samples with natural within-writer variability. Whether this significantly affected the results is uncertain and should be considered in the future. 

\subsubsection{EAA Collections}
Models trained on the EAA data set showed increased performance in the augmented condition compared to the non-augmented condition for Junclets and QuadHinge on test data. On the other hand, models for TCC, DeltaHinge, and Hinge showed a decreased performance in the augmented condition, and CoHinge showed no change in performance on test data. However, this is not reflected in the validation results (Table \ref{_data:DSS_cv}). Instead, validation results show a decrease in performance in the augmented condition for Junclets and TCC compared to the non-augmented condition, and an increase in performance for the remaining features. 

The results of the EAA data could be explained by the increase in standard deviations across all features for models trained on both augmented and non-augmented data compared to models trained on only non-augmented data. This increase indicates that models were less robust to new data in the augmented condition, which may have led to diverging test results. Additionally, the differences between test results and validation results within the conditions, e.g., QuadHinge, indicate overfitting. This likely follows from the small size of the data set.

A possible reason why models trained with the EAA data set were less robust in the augmented condition is that linear SVMs were inappropriate for the data. While they previously worked well for the Roman script on the MPS data set, temporal information in the features extracted from EAA may follow non-linear patterns. Data augmentation could have emphasized these non-linear patterns, making linear models too rigid. 

\subsection{Future research}
Scripts have different characteristics, possibly resulting in differing distributions of extracted features. Likewise, individual features capture varying attributes of handwriting. Therefore, temporal information on handwriting styles might follow different trends across various features. While linear SVMs performed well on the MPS data set for the features used in the current research, these potential differences in distributions were not considered. This could lead to a decrease in performance for models trained on augmented data. Hence, other kernels should be studied to obtain optimal models for individual features and scripts.

One of the risks with historical manuscript dating is that the majority of the samples from a period, or a year, originate from one writer. Rather than learning to distinguish between characteristics in handwriting styles specific to a particular period or year, models would learn traits specific to writers for these years. Data augmented to simulate variability between writers within time periods might lead to more robust models than when data is augmented to simulate a realistic within-writer variability. 

As mentioned in Section \ref{sec:lit_rev}, deep learning approaches outperformed statistical approaches on the MPS data set. Considering this, it would be interesting to investigate whether data augmentation might positively affect historical manuscript dating on smaller and heavier degraded manuscripts, such as the EAA collections. Moreover, using the shape evolution of individual characters with grapheme-based statistical features might bypass the issue of limited data and loss of information due to the resizing of images. 

\section*{\uppercase{Acknowledgement}}\label{sec:acknowledgement}
The study for this article collaborated with several research outcomes from the European Research Council (EU Horizon 2020) project: The Hands that Wrote the Bible: Digital Palaeography and Scribal Culture of the Dead Sea Scrolls (HandsandBible 640497), principal investigator: Mladen Popovi\'{c}. Furthermore, for the high-resolution, multi-spectral images of the Dead Sea Scrolls, we are grateful to the Israel Antiquities Authority (IAA), courtesy of the Leon Levy Dead Sea Scrolls Digital Library; photographer: Shai Halevi. Additionally, we express our gratitude to the Bodleian Libraries, University of Oxford, the Khalili collections, and the Staatliche Museen zu Berlin (photographer: Sandra Steib)  for the early Aramaic images. We also thank Petros Samara for collecting the Medieval Paleographical Scale (MPS) dataset for the Dutch NWO project. Finally, we thank the Center for Information Technology of the University of Groningen for their support and for providing access to the Peregrine high-performance computing cluster.

\bibliographystyle{plain}
{\small

}

\hfill
\section*{\uppercase{Appendix}}
\begin{table}[ht]
    \caption{The list of EAA images used in this research.}
    \label{apeen:list}
    \centering
        \begin{tabular}{lll}
        \hline
        A6\_11R & A6\_8    & NS\_A1r                                                    \\ \hline
        A6\_12R & B3\_1    & NS\_A2r                                                    \\ \hline
        A6\_13R & IA01     & NS\_A4r                                                    \\ \hline
        A6\_14  & IA03     & NS\_A5r                                                    \\ \hline
        A6\_15  & IA04     & NS\_A6r                                                    \\ \hline
        A6\_16  & IA06     & NS\_C1r                                                    \\ \hline
        A6\_3   & IA17     & NS\_C4r                                                    \\ \hline
        A6\_4   & IA21     & WDSP1\_1                                                   \\ \hline
        A6\_5   & Mur24\_1 & WDSP2                                                      \\ \hline
        A6\_7   & Mur24\_2 & \begin{tabular}[c]{@{}c@{}}Maresha\\ Ostracon\end{tabular} \\ \hline
        \end{tabular}
\end{table}

\end{document}